\documentclass[sigconf]{acmart}
\AtBeginDocument{%
  }
\usepackage{multirow}
\usepackage{subcaption}
\setcopyright{acmlicensed}
\copyrightyear{2018}
\acmYear{2018}
\acmDOI{XXXXXXX.XXXXXXX}
\acmConference[WWW '25]{Make sure to enter the correct
  conference title from your rights confirmation emai}{April 28--May 02,
  2025}{Sydney, Australia}

\acmISBN{978-1-4503-XXXX-X/18/06}

\begin{document}

\title{An archaeological Catalog Collection Method Based on Large Vision-Language Models}


\author{Honglin Pang\textsuperscript{1}, Yi Chang\textsuperscript{1}, Tianjing Duan\textsuperscript{2\textsubscript{*}}, Xi Yang\textsuperscript{1\textsubscript{*}}}
\affiliation{%
\institution{\textsuperscript{1}\institution{School of Artificial Intelligence, Jilin University, China} \country{}}
\institution{\textsuperscript{2}\institution{School of Archaeology, Jilin University, China} \country{}}
{\tiny\textsuperscript{*}Corresponding Author}
}





\renewcommand{\shortauthors}{Pang et al.}

\begin{abstract}
Archaeological catalogs, containing key elements such as artifact images, morphological descriptions, and excavation information, are essential for studying artifact evolution and cultural inheritance. 
These data are widely scattered across publications, requiring automated collection methods. However, existing Large Vision-Language Models (VLMs) and their derivative data collection methods face challenges in accurate image detection and modal matching when processing archaeological catalogs, making automated collection difficult. To address these issues, we propose a novel archaeological catalog collection method based on Large Vision-Language Models that follows an approach comprising three modules: document localization, block comprehension and block matching. Through practical data collection from the Dabagou and Miaozigou pottery catalogs and comparison experiments, we demonstrate the effectiveness of our approach, providing a reliable solution for automated collection of archaeological catalogs.
\end{abstract}




\maketitle

\begin{figure}
  \includegraphics[width=0.45\textwidth]{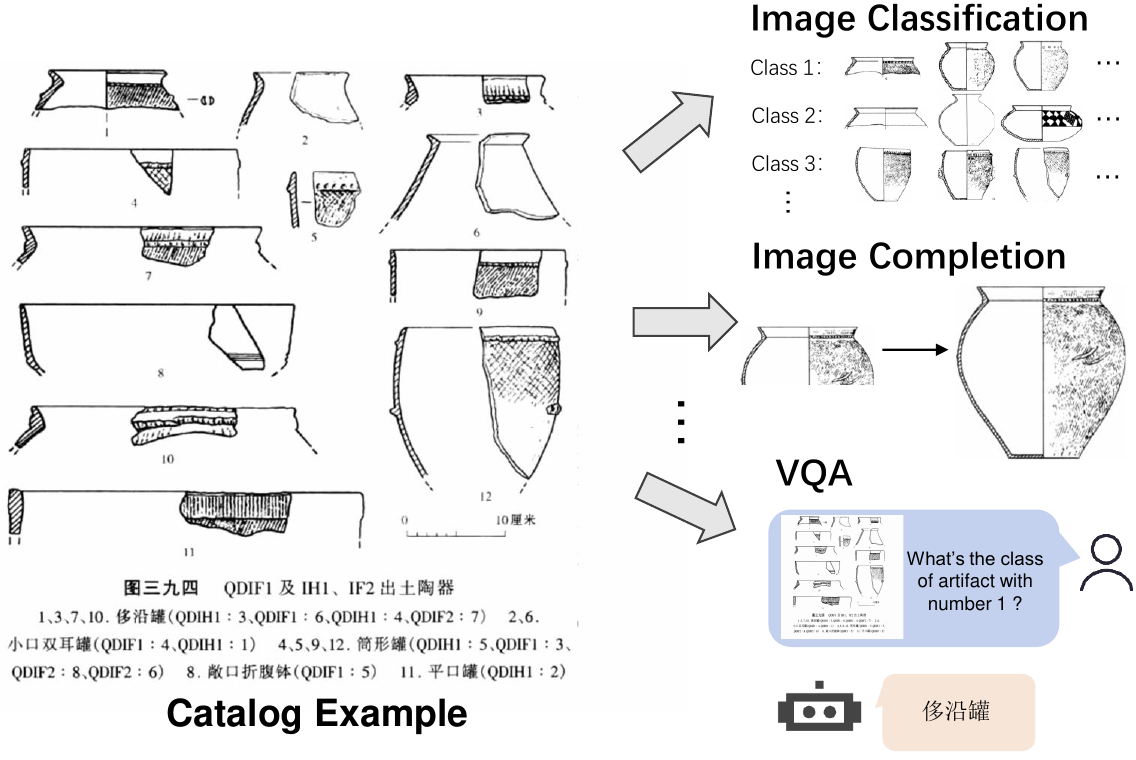}
  \caption{Target tasks for catalog datasets.}
  \Description{}
  \label{fig:tasks}
\end{figure}

\begin{figure*}
  \includegraphics[width=\textwidth]{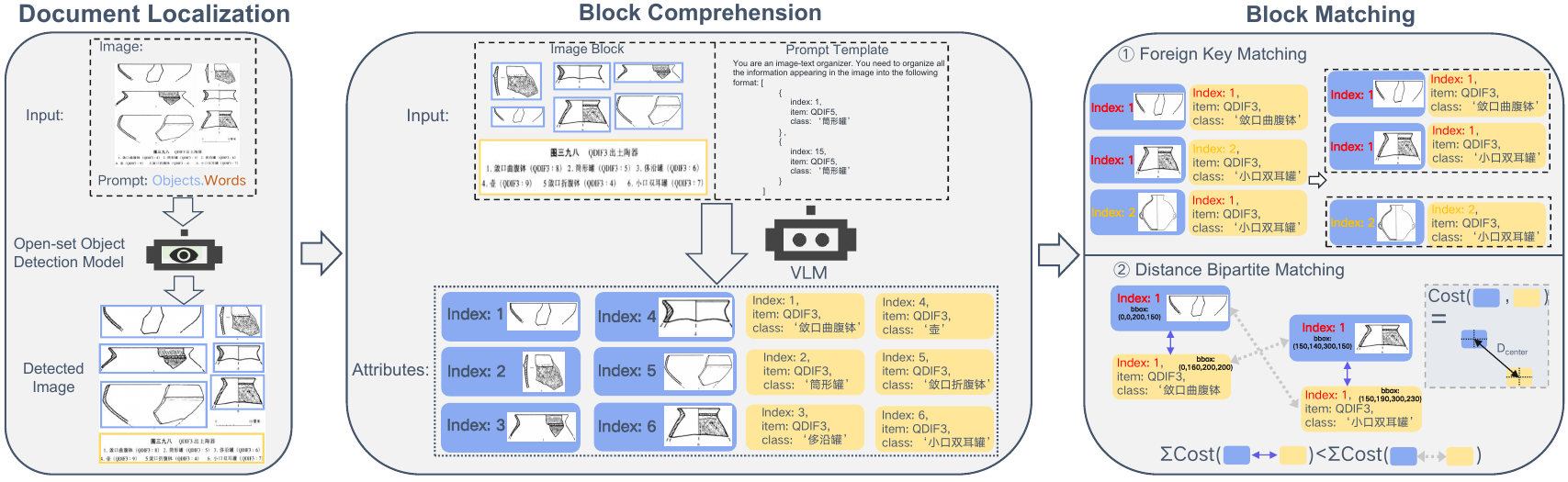}
  \caption{The pipeline consists of three main modules: Document Localization Module for segmenting image and text regions, Block Comprehension Module for converting visual and textual information into structured representations, and Block Matching Module for aligning multi-modal information through matching rules to achieve automated archaeological catalog collection.}
  \Description{}
  \label{fig:pipeline}
\end{figure*}

\section{Introduction}

Archaeological catalogs, containing crucial information such as artifact images, morphological descriptions, and excavation details, serve as fundamental resources for understanding artifact evolution patterns and cultural inheritance~\cite{pang2024pottery,zhou2023multi}. These archaeological datasets can be leveraged for various computational tasks, including classification tasks for artifact categorization, generative tasks for artifacts reconstruction, and visual question answering tasks, as shown in Figure \ref{fig:tasks}. However, these valuable data are widely dispersed across numerous archaeological publications, and Traditional manual collection methods are labor-intensive and error-prone, making it challenging to meet the growing demands of modern archaeological research.

Recent advances in Large Vision-Language Models (VLMs), have shown promising potential in automated data collection and processing. These models have demonstrated remarkable capabilities in understanding and analyzing both visual and textual information in various fields. However, when applied to archaeological catalogs, existing VLMs\cite{claude,bai2023qwen,openai2024gpt4technicalreport,zhang2022glipv2,liu2025grounding,ren2024grounding} and their derivative data collection methods\cite{yu2024visrag,perot2023lmdx,biswas2024robustness,wu2024structured,tang2024pdfchatannotator} encounter technical challenges in accurate image detection and modal matching, leading to collection failures.

To address these challenges, we propose a novel archaeological catalog collection method based on large vision-language models that follows an approach comprising three modules: document localization, block comprehension, and block matching. First, we employ open-set object detection models to localize and segment document blocks. Then, we process these document blocks for comprehension and describe them in terms of attributes. Finally, we implement matching rules based on foreign key linkages and bipartite graph matching to complete modal matching.

To validate the effectiveness of our proposed method, we conducted practical data collection experiments and comparison experiments using the Dabagou and Miaozigou pottery catalogs. Through these experiments, we demonstrated that our approach significantly improves the accuracy of automated archaeological catalog collection. The results of our comparison experiments highlight the reliability of our method in overcoming the limitations of existing VLM-based data collection methods.

\section{Related Work}
Recent advances in VLMs have enabled significant progress in data collection, particularly in two key aspects: key information extraction and relationship extraction. 

\textbf{Key Information Extraction Based on VLM.}
Existing work can extract key information from documents \cite{yu2024visrag,perot2023lmdx,biswas2024robustness,wu2024structured} and reorganize it into related claims \cite{chen2024unified}. This key information can be extracted from both text and images. Sometimes, the key information is in the form of images, and current methods \cite{li2022grounded,zhang2022glipv2,liu2025grounding,ren2024grounding} can directly ground the information in images. However, for specialized data, direct grounding remains challenging. Particularly when images contain complex specialized information, existing grounding methods may not accurately identify and extract this information.

\textbf{Relationship Extraction Based on VLM.}
Recent methods have demonstrated the ability to leverage Vision-Language Models (VLMs) to extract relationships between entities and construct knowledge graphs \cite{he2023using}. LMDX \cite{perot2023lmdx} utilizes these relationships for binding, but it is limited to a single modality. On the other hand, PDFChatAnnotator \cite{tang2024pdfchatannotator} connects knowledge across different modalities within documents. However, the rules for modality binding in this method remain relatively simple, leading to potential issues in complex scenarios.

\section{Method}

\subsection{Overview}
To address the challenges in archaeological catalog collection, we propose a pipeline consisting of three main modules: Document Localization Module, Block Comprehension Module, and Block Matching Module (as illustrated in Figure \ref{fig:pipeline}). Document Localization Module first segments the input catalogs into distinct regions, separating images from their corresponding annotations. Subsequently, Block Comprehension Module processes both visual and textual information, converting them into structured textual attributes for storage. Finally, Block Matching Module employs matching algorithms to align information from different modalities, thereby completing the data collection process.
\subsection{Document Localization Module}
To accurately locate images and annotations in archaeological catalogs, we leverage an open-set object detection model with specifically designed prompts. Given an input catalog page $I$, we formulate the detection task as:

\begin{equation}
    B = f(I, P_{img}, P_{text})
\end{equation}

where $B$ represents the detected boxes, $f$ is the detection model, and $P_{img}$, $P_{text}$ denote the prompts designed for detecting image blocks and text blocks respectively. The model outputs two sets of bounding boxes $B = \{B_{img}, B_{text}\}$, where $B_{img} = \{b_1^i, b_2^i, ..., b_n^i\}$ represents image blocks and $B_{text} = \{b_1^t, b_2^t, ..., b_m^t\}$ represents text blocks. Each bounding box $b_k = (x_c, y_c, w, h)$ specifies the location and size of the detected region.

\subsection{Block Comprehension Module}
For each detected region, we employ a VLM to generate structured attributes. Given the set of bounding boxes $B$ from the detection module, the comprehension process can be formalized as:
\begin{equation}
C = g(B, P_c)
\end{equation}
where $C = \{C_{img}, C_{text}\}$ represents the comprehension results, consisting of image region comprehension results $C_{img} = \{c_1^i, c_2^i, ..., c_n^i\}$ and text region comprehension results $C_{text} = \{c_1^t, c_2^t, ..., c_m^t\}$. $g$ is the VLM, and $P_c$ is the prompt template guiding the description format.
For each region $b_q \in B$, its comprehension result $c_q$ is a dictionary containing key attributes:
\begin{equation}
c_q = \{(k_1:v_1), (k_2:v_2), ..., (k_p:v_p)\}
\end{equation}
where $k_j$ represents attribute keys and $v_j$ represents corresponding values. 

\subsection{Block Matching Module}
To establish correspondences between image blocks and text blocks, we propose a two-stage matching strategy. 
\begin{figure}
  \includegraphics[width=0.45\textwidth]{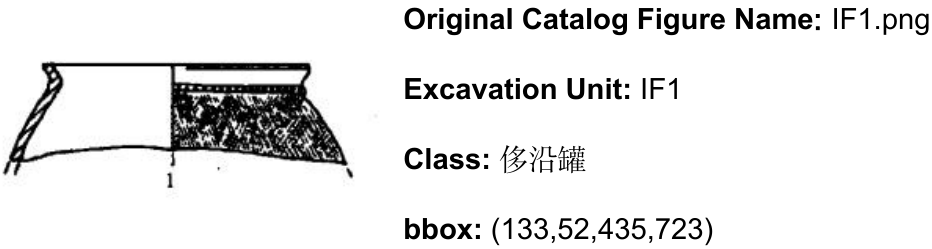}
  \caption{Annotations of a pottery example.}
  \Description{}
  \label{fig:annotation}
\end{figure}

\textbf{Foreign Key Matching.}
For each element $c_q$ in image region comprehension results $C_{img}$ and text region comprehension results $C_{text}$, we define its foreign key set $FK_q$ as:
\begin{equation}
FK_q = \{v_j | k_j \in K_{foreign}\}
\end{equation}
where $K_{foreign}$ is the predefined set of foreign key attributes. The matching degree between two blocks $c_a$ and $c_b$ can be measured by the overlap of their foreign key sets:
\begin{equation}
M(c_a, c_b) = \frac{|FK_a \cap FK_b|}{|FK_a \cup FK_b|}
\end{equation}
When $M(c_a, c_b)=1$, indicating complete foreign key matching between two blocks, we consider these blocks to be corresponding. This can be represented as a set of matching pairs:
\begin{equation}
Matches_{FK} = \{(c_a, c_b) | c_a \in C_{img}, c_b \in C_{text}, M(c_a, c_b)=1\}
\end{equation}

\begin{figure*}[htbp]
    \centering
    \begin{subfigure}[t]{0.48\textwidth}
        \centering
        \includegraphics[width=\linewidth]{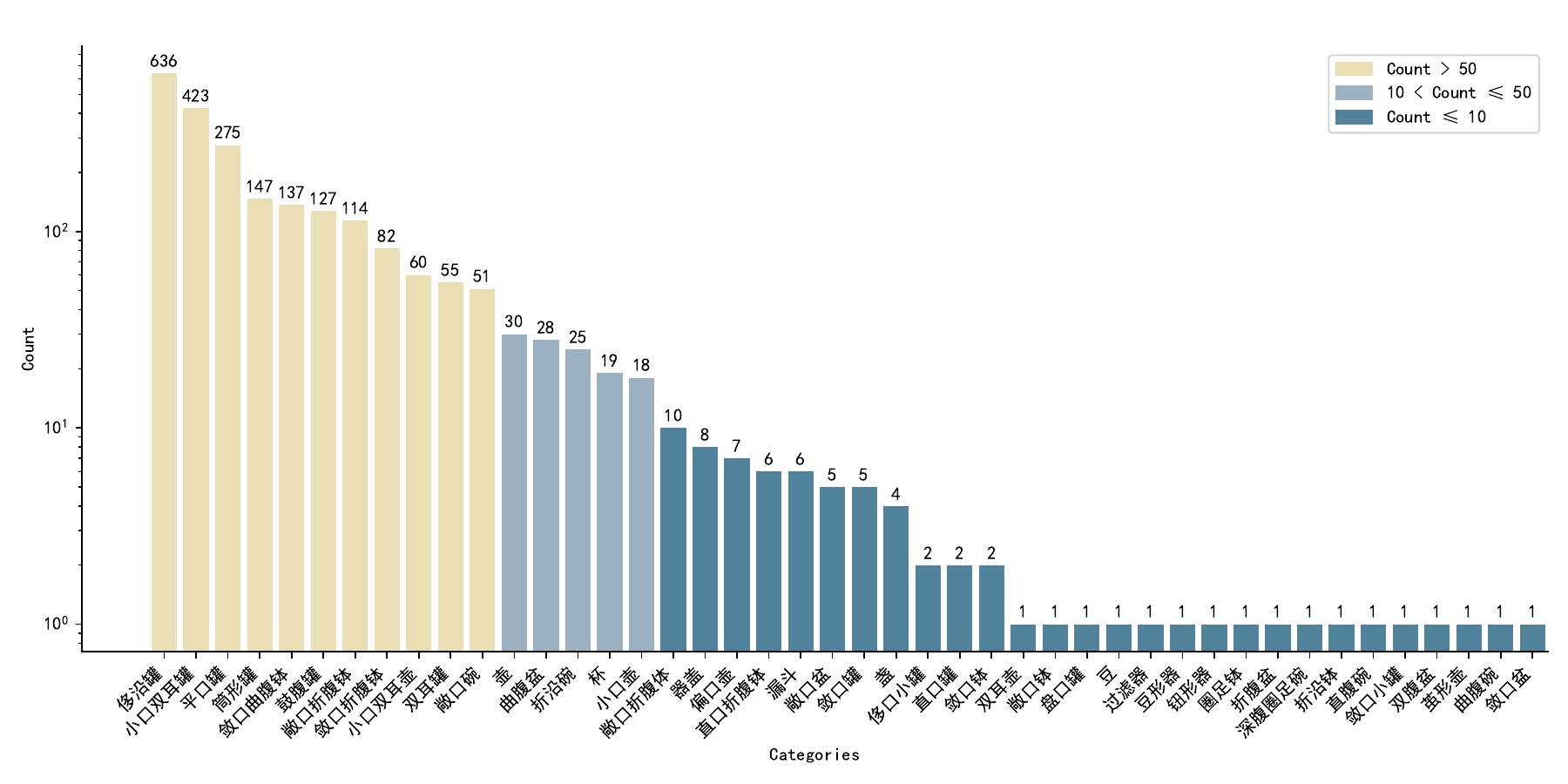}
        \caption{Category Distribution of the Dataset}
        \label{fig:subfig1}
    \end{subfigure}
    \hfill
    \begin{subfigure}[t]{0.48\textwidth}
        \centering
        \includegraphics[width=\linewidth]{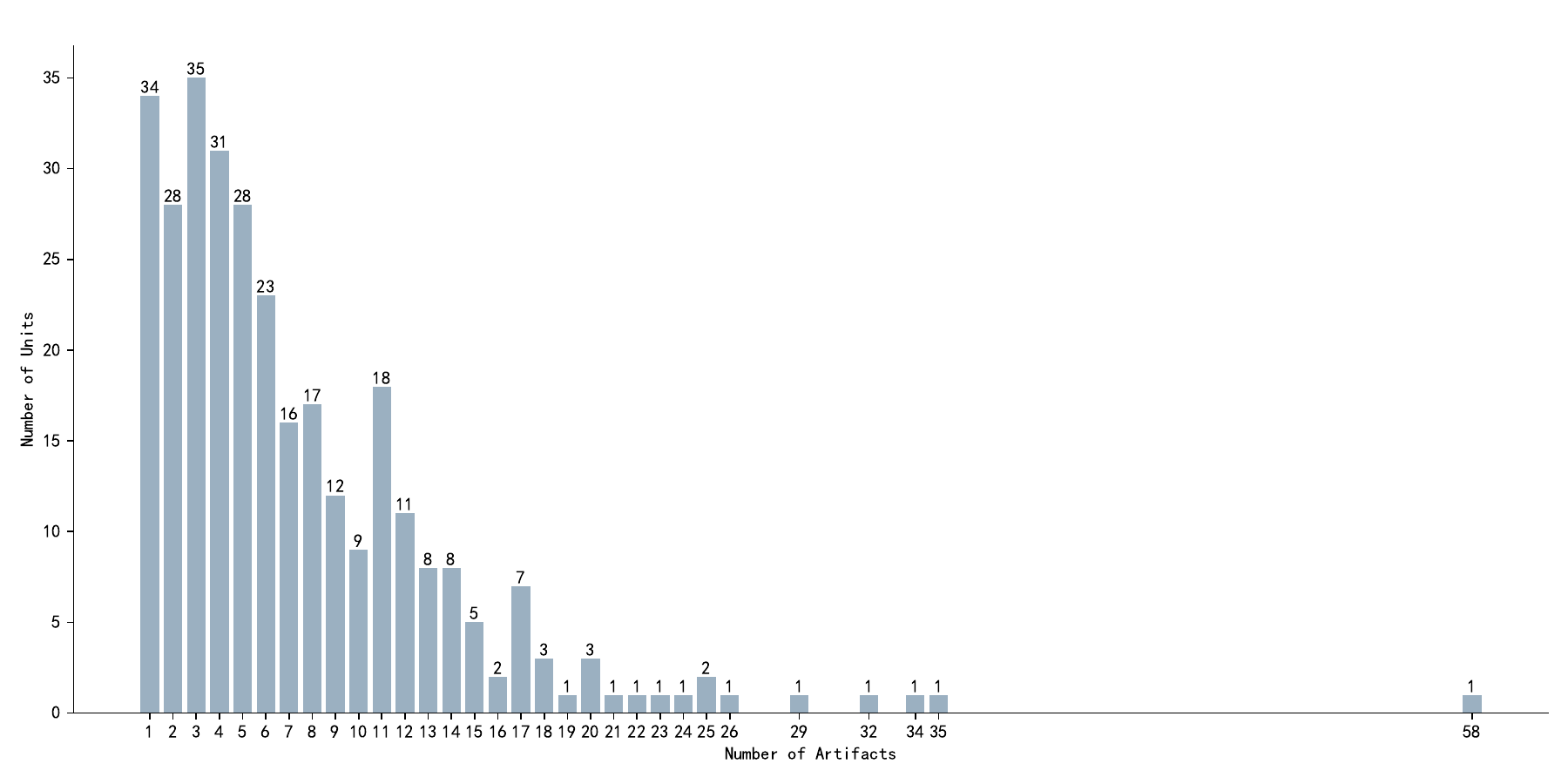}
        \caption{Statistical Distribution of Artifacts across Units}
        \label{fig:subfig2}
    \end{subfigure}
    \caption{Statistical analysis of the dataset: (a) shows the distribution of samples across different categories, and (b) illustrates the distribution of artifacts per unit.}
    \Description{The figure presents two distributions: the first showing category-wise sample distribution and the second showing the number of artifacts per unit in the dataset.}
    \label{fig:statistics}
\end{figure*}

\textbf{Distance Bipartite Matching.}
Multiple image regions may share identical foreign keys, especially when they belong to the same logical group or represent related information. For example, several product images might reference the same product ID.
Therefore, for many-to-many matching cases in $Matches_{FK}$, we further refine the correspondences using bipartite matching based on spatial distance. Given a group of matched blocks $G = \{(c_i^{img}, c_j^{text})\} \subset G_k$ where $G_k$ represents one many-to-many matching group, we construct a bipartite graph where nodes represent image blocks and text blocks respectively.

The cost matrix $D$ for bipartite matching is defined using the Euclidean distance between block centers. We then solve the minimum-weight bipartite matching problem. The final block correspondences are obtained by combining both matching stages.The process can be expressed using the following formula: 
\begin{equation}
D_{ij} = \|center(b_i^{img}) - center(b_j^{text})\|_2
\end{equation}

\begin{equation}
\mathcal{M}^* ={\arg\min}\,\sum_{(i,j) \in G_k} D_{ij}
\end{equation}

\begin{equation}
Matches_{final} = (Matches_{FK} \setminus \bigcup_{k} G_k) \cup \mathcal{M}^*
\end{equation}
where $center(\cdot)$ returns the center coordinates of a block's bounding box, $\mathcal{M}_k^*$ is the optimal bipartite matching for group $G_k$. This process preserves one-to-one matches from the foreign key stage while resolving many-to-many cases through distance-based bipartite matching.

\section{Experiments}



\subsection{Data Collection and Statistics.}
The pottery data collection from Miaozigou and Dabagou sites is crucial for archaeological research, as these sites represent the significant Miaozigou Culture. This data provides essential evidence for understanding the unique developmental trajectory of Neolithic cultures in central and southern Inner Mongolia.

Using the proposed method, we collected 302 pottery catalog PDF pages from Dabagou and Miaozigou sites. With minimal manual correction, we obtained 2,301 data pairs, demonstrating the effectiveness of our approach.


\textbf{Annotations.}
We compiled the collected attributes, location information and file information into annotations. For each sample, we recorded the following information, as Figure~\ref{fig:annotation} shown:
original catalog figure name, pottery image, Excavation unit, morphological class and bounding box coordinates.


\textbf{Statistics.}
We conducted statistical analysis on the dataset as shown in Figure \ref{fig:statistics}. The dataset contains 44 classes. As can be observed, the distribution across shape classes exhibits a long-tail pattern, with significant class imbalance. Additionally, we analyzed the number of artifacts per excavation unit, as presented in Table \ref{tab:performance_comparison}. These artifacts were excavated from 310 different units, with most units containing fewer than 10 artifacts.




\begin{table*}[htbp]
\centering
\caption{Performance comparison of our method using different VLMs against baseline methods}
\label{tab:performance_comparison}
\begin{tabular}{l|ccc|cccc}
\hline
\multirow{3}{*}{Method} & \multicolumn{3}{c|}{\textit{Ours (with different VLMs)}} & \multicolumn{4}{c}{\textit{Baseline Methods}} \\
\cline{2-8}
& GPT-4o & \begin{tabular}[c]{@{}c@{}}Claude 3.5\\Sonnet\end{tabular} & Qwen-VL & \begin{tabular}[c]{@{}c@{}}GPT-4o\\only\end{tabular} & \begin{tabular}[c]{@{}c@{}}Claude 3.5\\Sonnet only\end{tabular} & \begin{tabular}[c]{@{}c@{}}Qwen-VL\\only\end{tabular} & \begin{tabular}[c]{@{}c@{}}Claude 3.5 Sonnet +\\Grounding-DINO\end{tabular} \\
\hline
AP (\%) & 32.8 & \textbf{35.4} & 27.4 & - & - & 1.6 & 0 \\
\hline
\end{tabular}
\end{table*}

\subsection{Comparison Experiments}
We use the cleaned dataset as ground truth to calculate metrics for results obtained from each method, in order to evaluate their effectiveness.

\textbf{Metrics.}
We evaluate our method using Average Precision (AP) with an IoU threshold of 90\%. The high IoU threshold ensures strict evaluation of localization accuracy. 

\textbf{Comparison between different VLMs.}
To demonstrate the generalizability of our pipeline across different vision-language models, we conduct experiments with GPT-4o\cite{openai2024gpt4technicalreport}, Claude 3.5\cite{claude} Sonnet, and Qwen-VL\cite{bai2023qwen}. 


As shown in Table \ref{tab:performance_comparison}, Claude 3.5 Sonnet achieves the best performance with an AP of 35.4\%, outperforming other VLMs. This demonstrates that while our pipeline is model-agnostic, the choice of VLM can impact the overall performance.

\textbf{Comparison with baseline.}
We compare our method with two baseline methods: (1) using GPT 4o, Claude 3.5 Sonnet and Qwen-VL alone for detection and comprehension (2) combining Claude 3.5 Sonnet with open-set object detection. 
In both (1) and (2), we first use a VLM to analyze catalog numbers, excavation units, and other information. In (1), we then use the same VLM model to ground the artifact images, while in (2), we apply Grounding-DINO model for grounding the images.
The results are presented in Table \ref{tab:performance_comparison}.


Our experimental results validate the effectiveness of the proposed pipeline. While GPT-4o and Claude 3.5 Sonnet lack object detection capabilities and thus could not complete the catalog collection task, our method achieved a 33.8\% improvement in AP compared to using Qwen-VL alone. Although large language models can leverage existing object detection models, these detectors' limited semantic understanding restricts their task performance. In contrast, our approach successfully completes the task and outperforms baseline models, demonstrating its superiority.

\section{Conclusion}
In this paper, we presented a novel approach for archaeological catalog collection utilizing Large Vision-Language Models. Our method addresses the challenges in automated archaeological data processing through localization, comprehension and modal matching.

Experimental validation using the Dabagou and Miaozigou pottery catalogs demonstrated the effectiveness and reliability of our approach. The results show significant improvements in automated archaeological catalog collection accuracy compared to existing VLM-based methods. Our proposed framework provides a robust solution for processing archaeological documentation.

Future work could focus on extending this approach to diverse archaeological artifacts and exploring additional matching algorithms to further enhance the system's performance.


\bibliographystyle{ACM-Reference-Format}
\bibliography{sample-base}


\end{document}